\documentclass[11pt,a4paper]{article}

\usepackage[utf8]{inputenc}
\usepackage[T1]{fontenc}
\usepackage{microtype}
\usepackage{amsmath,amssymb}
\usepackage[a4paper,margin=2.5cm]{geometry}
\usepackage{graphicx}
\usepackage{array}
\usepackage{tabularx}
\usepackage{booktabs}
\usepackage{multirow}
\usepackage{threeparttable}
\usepackage{lscape}
\usepackage{float}
\usepackage{listings}
\usepackage{inconsolata}
\usepackage{tikz}
\usetikzlibrary{positioning,arrows.meta,shapes.geometric,shapes.misc,fit,backgrounds,calc,decorations.pathreplacing}
\usepackage[numbers,sort&compress]{natbib}
\usepackage[colorlinks=true,linkcolor=blue,citecolor=blue,urlcolor=blue]{hyperref}

\newcommand{\tightcols}{\setlength{\tabcolsep}{4pt}}

\lstdefinelanguage{cypher}{
  keywords={MATCH, RETURN, WHERE, CREATE, MERGE, DELETE, SET, WITH, OPTIONAL, AS, AND, OR, NOT, IN, IS, NULL, ORDER, BY, LIMIT, SKIP, UNION, CALL, YIELD, UNWIND, FOREACH, EXPLAIN, PROFILE, COUNT, SUM, AVG, MIN, MAX},
  keywordstyle=\bfseries,
  comment=[l]{//},
  morestring=[b]'',
  sensitive=true,
}

\definecolor{lstbg}{HTML}{F5F5F5}

\title{CYGNET: Cypher Gate for Neural Execution Triage and Cost Containment}

\author{Nikodem Tomczak \\ Thulge Labs, Singapore}

\date{\today}

\begin{document}

\maketitle

\begin{abstract}
Large language models acting as agents over knowledge graphs generate Cypher queries that fail structurally, crashing at the database and consuming compute without returning any result, or semantically, where valid Cypher executes but returns wrong results. Existing tooling does not distinguish the two failure modes and existing benchmarks aggregate both into a single accuracy metric. We show that placing a pre-execution gate between a query-generating system and a production graph database catches structural failures before they reach production. The gate validates query structure through a four-backend chain that culminates in execution against a mirror graph, a minimal synthetic construction built from the schema, at a median latency of 5.6 ms per query. Structurally broken queries are routed to a corrector that feeds structured error descriptions to a language model and iterates until the query passes validation or an attempt budget is exhausted. On seven schemas from the CypherBench benchmark the validation and correction pipeline maintains generation accuracy on every model tested, confirming that it operates as a safe defensive layer without introducing semantic regressions. The corrector achieves 81\% to 95\% success at repairing structurally broken queries across five language models, with a mean of 89\%. To our knowledge no public Cypher corpus exists with categorical structural error labels. We generate one covering four error categories across nine schemas and evaluate against it. The gate catches 100\% of parse errors, 100\% of constraint violations, and 100\% of schema-reference errors in path queries with labelled endpoints, at a false-positive rate of zero across 1135 queries. Schema swaps in standalone queries without relationships are caught at 48\% through cascading property checks. Property sibling-swaps where the substituted name is valid on the target label are structurally undetectable and score 0\%, marking the formal boundary where structural validation ends and semantic validation must take over. A planner-based cost gate identifies catastrophic plan structures such as unbounded variable-length paths before execution.
\end{abstract}

\section{Introduction}
\label{sec:introduction}

Knowledge graphs underpin a growing share of retrieval-augmented generation systems, scientific data management platforms, and enterprise search infrastructure \citep{zhu2025kg2rag,peng2024kgrag_survey,edge2024graphrag}. Their structured representation of entities and relationships provides factual grounding that flat text retrieval cannot, and their dynamic update semantics suit operational contexts in which the underlying knowledge changes between queries. Large language models operating as agents over these graphs translate natural-language questions into queries in graph query languages such as Cypher \citep{francis2018cypher} and execute them against production graph databases. Conversational and multi-agent systems including Chatty-KG \citep{omar2025chattykg} and AGENTiGraph \citep{zhao2025agentigraph} demonstrate the architectural pattern in which one or more language-model agents construct, refine, and execute graph queries inside a loop.

Agent-generated queries fail in two fundamentally different ways. A structural failure occurs when the query is syntactically malformed, references a label or relationship type absent from the schema, passes a value of the wrong type to a property, or violates a declared constraint. A structurally broken query crashes at the database, consumes compute for parsing and planning, and returns no useful result. A semantic failure occurs when the query is structurally valid Cypher that executes without error but returns wrong results because it does not correctly express the user's intent. Structural failures are detectable before execution by inspecting the query against the known schema.

A pre-execution gate that catches structural failures prevents wasted compute, reduces latency on the failure path, and gives the generating system actionable feedback that a post-hoc database error does not. Every query that would crash at the production database consumes parsing and planning resources before the database reports the error. On graphs where query execution is metered such as cloud-hosted databases and shared-resource clusters, every failed query is a cost event with zero return. A structural gate that catches these queries at a low median latency eliminates the wasted compute and returns structured feedback in a fraction of the time the database would have taken to fail. CyVer \citep{mandilara2025cyver} validates Cypher syntax, schema references, and property access against the database schema with structured error reporting. LangChain's \texttt{CypherQueryCorrector} \citep{langchain2023corrector} repairs relationship-direction errors. Production Model Context Protocol (MCP) servers including Neo4j Labs' \texttt{mcp-neo4j-cypher} \citep{neo4j2024mcp} expose schema introspection and post-hoc execution controls such as timeouts and truncation. Ozsoy \citep{ozsoy2026confidence} extends confidence-based Text2Cypher inference with grammar and schema-constraint filtering as a post-generation filter. SynthCypher \citep{tiwari2024synthcypher} creates synthetic Neo4j databases populated with LLM-generated data to enable execution-based verification of generated training data. SyntheT2C \citep{zhong2025synthet2c} validates generated Cypher by executing against existing medical knowledge graph databases. Both use execution-based verification for offline training-data generation rather than runtime pre-execution validation. None of these systems combine pre-execution structural validation, planner-based cost gating, and structured refinement feedback into an integrated runtime gate for agent-generated Cypher.

In this contribution we target structural failures in agent-generated Cypher. We implement a four-backend validator chain composed of a regex-based fast filter, an ANTLR-based \citep{parr2012antlr} abstract syntax tree (AST) parser, a Neo4j EXPLAIN backend, and a mirror-execute backend that runs queries inside rolled-back transactions against a synthetic mirror graph. The chain catches four categories of structural error (parse errors, schema-reference errors, property-type errors, and constraint violations) and reports each with a structured error payload that downstream systems can act on. An EXPLAIN-based cost gate parses planner output to identify expensive query patterns before execution. Five pluggable correctors implementing a single-shot protocol, combined with a generic refinement loop, route structured errors to a language model for iterative repair of structurally broken queries. Validation and correction run entirely against the mirror graph at a median latency of 5.6 ms, with no contact to the production database until the query passes structural validation. To our knowledge no public Cypher corpus provides categorical structural error labels. We generate a template-based corpus covering four error categories across nine schemas of varying shape and evaluate against it. The gate catches 100\% of parse errors, 100\% of constraint violations, and 100\% of schema-reference errors in path queries with labelled endpoints, at a false-positive rate of zero across 1135 queries. Schema swaps in standalone queries are caught at 48\% through cascading property checks, and property sibling-swaps where the substituted name is valid on the target label are structurally undetectable. On an external corpus of 683 independently authored valid queries the chain flagged 39, all valid Cypher 5+ syntax that the bundled Cypher 9 grammar does not parse. Across five language models and five corrector implementations, the best corrector achieves 81\% to 95\% success at repairing structurally broken queries after up to three refinement attempts, with a mean of 89\%. On seven CypherBench schemas (2348 questions) the pipeline does not degrade generation accuracy on any model, confirming it operates as a safe defensive layer.

\section{Methods}
\label{sec:methods}

\subsection{Architecture and schema sources}
\label{sec:methods-architecture}

We place a programmable gate between a Cypher-generating system and a Neo4j graph database. The gate validates the structure of an incoming query against a known schema, estimates the query's execution cost, and returns either a permission to execute or a structured error description suitable for downstream refinement. The gate targets systems where a language model generates queries inside an agent loop, where production operators want to bound the cost their database absorbs from malformed or runaway queries, and where the cost of an unintended execution against a large graph exceeds the cost of running the gate by several orders of magnitude.

The gate is accessible through three transports, a direct in-process API, a Model Context Protocol (MCP) server for MCP-speaking hosts, and an HTTP server for language-agnostic services. All three share the same internal core described in the remainder of this section.

\begin{figure}[h]
\centering
\begin{tikzpicture}[>=Stealth, font=\sffamily\small, scale=0.62, transform shape]

\definecolor{cygblue}{HTML}{3B82F6}
\definecolor{cyggreen}{HTML}{22C55E}
\definecolor{cygred}{HTML}{EF4444}
\definecolor{cyggrey}{HTML}{94A3B8}
\definecolor{cyglblue}{HTML}{EFF6FF}

\tikzset{
  gatebox/.style={rounded corners=3pt, draw=cygblue, fill=white, line width=0.8pt,
    minimum height=0.9cm, minimum width=2.2cm, inner sep=4pt, align=center,
    font=\sffamily\small},
  extbox/.style={rounded corners=3pt, draw=cyggrey, fill=cyggrey!20, line width=0.8pt,
    minimum height=1cm, minimum width=2.2cm, align=center, font=\sffamily\small},
  dbshape/.style={rounded corners=3pt, draw=cyggrey, fill=cyggrey!20, line width=0.8pt,
    minimum height=1.2cm, minimum width=2cm, align=center, font=\sffamily\small},
  mirrorshape/.style={rounded corners=3pt, draw=cygblue, fill=cygblue!10, line width=0.8pt,
    minimum height=1cm, minimum width=1.6cm, align=center, font=\sffamily\footnotesize},
  passarr/.style={->, draw=cyggreen, line width=1.1pt},
  failarr/.style={->, draw=cygred, line width=0.9pt},
  dataarr/.style={->, draw=black!65, line width=0.9pt},
  mirrorline/.style={dashed, draw=cygblue, line width=0.7pt},
  retryarr/.style={->, dashed, draw=cygblue, line width=1pt},
  elab/.style={font=\sffamily\scriptsize, inner sep=1.5pt, fill=white, fill opacity=0.85, text opacity=1},
}

\fill[cyglblue, rounded corners=6pt] (3.2, 0.4) rectangle (19.2, 7.8);
\draw[dashed, cygblue, line width=0.9pt, rounded corners=6pt] (3.2, 0.4) rectangle (19.2, 7.8);
\node[anchor=north west, text=cygblue, font=\sffamily\bfseries\small] at (3.4, 7.7) {Gate};

\node[extbox] (llm)  at (1.0, 6.5) {LLM\\Agent};
\node[dbshape] (prod) at (21.5, 4.5) {Production\\Neo4j};

\node[gatebox] (builtin) at (5.2,  6.5) {Regex};
\node[gatebox] (ast)     at (8.2,  6.5) {AST};
\node[gatebox] (explain) at (11.2, 6.5) {EXPLAIN};
\node[gatebox, minimum width=2.6cm] (mexec) at (14.8, 6.5) {Mirror-Execute};

\node[mirrorshape] (mneo) at (13.0, 4.5) {Mirror\\Neo4j};

\node[gatebox] (cost) at (17.4, 4.5) {Cost Gate};

\node[gatebox, minimum width=12.5cm] (corr) at (11.2, 1.5) {Corrector};


\draw[dataarr] (llm.east) -- node[above, elab] {Cypher query} (builtin.west);

\draw[passarr] (builtin.east) -- node[above, elab, text=cyggreen] {pass} (ast.west);
\draw[passarr] (ast.east)     -- node[above, elab, text=cyggreen] {pass} (explain.west);
\draw[passarr] (explain.east) -- node[above, elab, text=cyggreen] {pass} (mexec.west);

\draw[mirrorline] (explain.south) -- node[left, elab, text=cygblue, pos=0.4] {plan} (mneo.north west);
\draw[mirrorline] (mexec.south) -- node[right, elab, text=cygblue, pos=0.4] {execute} (mneo.north east);

\draw[passarr] (mexec.south east) -- node[midway, above, sloped, elab, text=cyggreen] {pass} (cost.north west);

\draw[passarr] (cost.east) -- node[above, midway, elab, text=cyggreen] {validated} (prod.west);

\draw[failarr] (builtin.south) -- (builtin |- corr.north);
\draw[failarr] (ast.south) -- (ast |- corr.north);
\draw[failarr] (explain.south) -- (explain |- corr.north);
\draw[failarr] (mexec.south) -- (mexec |- corr.north);

\node[elab, text=cygred] at ([xshift=-8pt]builtin |- {0,4.0}) {fail};

\draw[failarr] (cost.south) -- (cost |- corr.north);
\node[elab, text=cygred] at ([xshift=8pt]cost |- {0,3.0}) {fail};

\draw[retryarr] (corr.west) .. controls (3.6, 1.5) and (3.6, 6.0) ..
  node[left, midway, elab, text=cygblue, align=center] {retry\\(up to 3)} (builtin.south west);

\draw[dataarr] (prod.south) -- (21.5, -0.3) --
  node[above, midway, elab] {results} (1.0, -0.3) -- (llm.south);

\end{tikzpicture}
\caption{Gate architecture and validation pipeline. Generated Cypher enters the four-backend validator chain (left to right). Any failure routes to the corrector with a structured error payload. The corrector retries through the full chain up to three times. Validated queries pass through the cost gate before reaching production.}
\label{fig:architecture}
\end{figure}
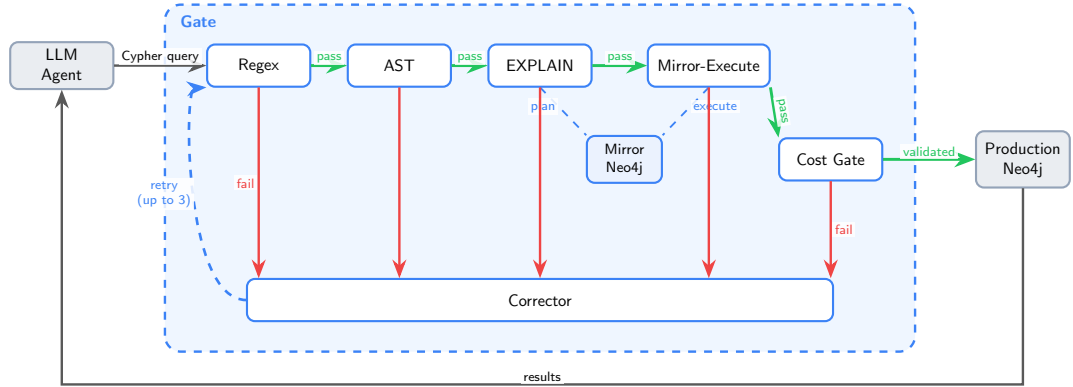

The gate operates against a schema description that names the labels, relationship types, properties, and constraints present in a target graph. The gate accepts the description from three sources, all resolving to the same internal representation. A YAML or JSON specification is appropriate when a graph's schema is well-known and stable, or when the gate operates in a development environment where a production database is not available. In-memory construction allows a caller with an existing schema object to pass it directly, which is appropriate when the schema is generated programmatically as part of a larger pipeline. Introspection of a live Neo4j instance is the third option. The gate connects to a database driver and runs schema-discovery queries. Where the APOC plugin library (a widely deployed Neo4j extension providing additional procedures and functions) is available the gate calls a single procedure that returns the full schema as a structured map. Where APOC is absent the gate falls back to built-in queries that enumerate constraints, indexes, node-type properties, and relationship-type properties individually. The fallback is automatic with a logged warning. Property sparsity (the fraction of nodes of a given label that carry a given property) is calculated by running a per-label count query, marking the property as sparse when its observed frequency falls below a configurable threshold (default 0.5). The gate supports a refresh operation that re-runs the schema load and atomically replaces the schema in the active chain.

\subsection{Validator backends}

The gate validates queries through a chain of four backends, each implementing a different strategy for detecting structural errors. A backend accepts a Cypher query and a schema and returns a result describing whether the query is structurally acceptable and, if not, which error category it falls into.

The regex backend applies regular expression patterns to extract labels, relationship types, and property accesses from a query, validating these against the schema's vocabulary and producing did-you-mean suggestions based on string similarity. It produces results in sub-millisecond time. Its limitations are inherent in regex parsing, and subqueries, pattern comprehensions, and procedure calls are recognised only partially.

The AST backend parses the query into an abstract syntax tree using a Cypher 9 grammar \citep{parr2012antlr}. The tree walk extracts label references, relationship references, and property accesses, including labels nested inside subqueries and properties accessed through complex patterns that the regex layer cannot reach. The AST backend also performs property type checking against the schema's declared types when literal values appear in WHERE clauses, and detects constraint violations on CREATE and MERGE patterns where literal values would conflict with declared existence or uniqueness constraints.

The EXPLAIN backend sends a planner-only request to a Neo4j driver and maps any returned exceptions or notifications to the error vocabulary. The Neo4j planner is authoritative for syntax, schema reference, property existence on schema-bound variables, and certain forms of constraint analysis. The EXPLAIN backend catches references the regex and AST backends cannot, including those inside Cypher syntax the AST grammar does not cover. Its limitation is that planning does not execute the query and therefore does not detect runtime-only errors such as missing procedures, parameter-type mismatches, scope errors in projections, and certain forms of implicit type coercion.

The mirror-execute backend executes the query inside a transaction against a mirror Neo4j instance (described in the next subsection) and rolls the transaction back unconditionally. Exceptions from the execution attempt are caught and mapped to the error vocabulary, covering syntax errors, semantic errors, missing procedures, missing parameters, and type errors. Writes performed inside the transaction are unwound by the rollback so the mirror remains structurally stable across calls.

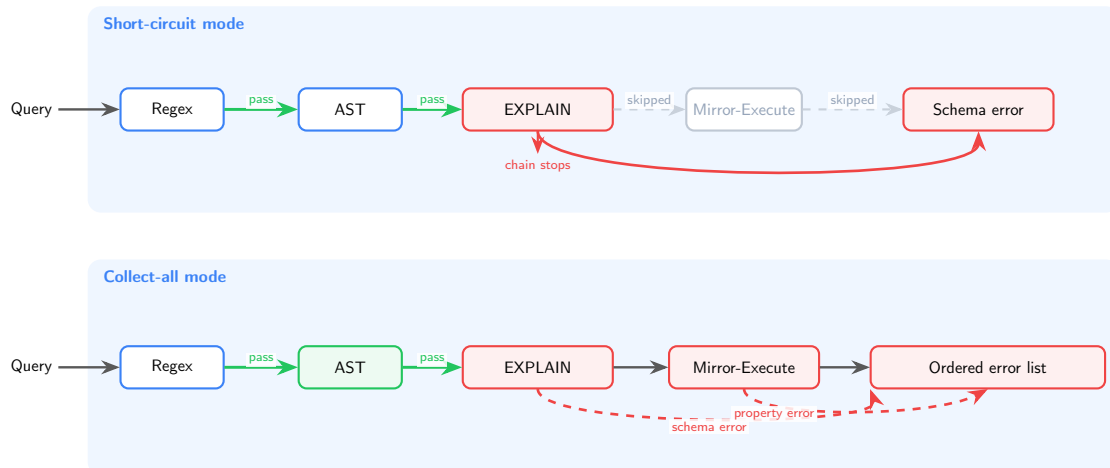
\begin{figure}[h]
\centering
\begin{tikzpicture}[>=Stealth, font=\sffamily\small, scale=0.62, transform shape]

\definecolor{cygblue}{HTML}{3B82F6}
\definecolor{cyggreen}{HTML}{22C55E}
\definecolor{cygred}{HTML}{EF4444}
\definecolor{cyggrey}{HTML}{94A3B8}
\definecolor{cyglblue}{HTML}{EFF6FF}

\tikzset{
  vbox/.style={rounded corners=3pt, draw=cygblue, fill=white, line width=0.8pt,
    minimum height=0.9cm, minimum width=2.2cm, inner sep=4pt, align=center,
    font=\sffamily\small},
  resultbox/.style={rounded corners=3pt, draw=cygred, fill=cygred!8, line width=0.8pt,
    minimum height=0.9cm, minimum width=3.2cm, inner sep=4pt, align=center,
    font=\sffamily\small},
  passbox/.style={rounded corners=3pt, draw=cyggreen, fill=cyggreen!8, line width=0.8pt,
    minimum height=0.9cm, minimum width=2.2cm, inner sep=4pt, align=center,
    font=\sffamily\small},
  skipbox/.style={rounded corners=3pt, draw=cyggrey!60, fill=white, line width=0.8pt,
    minimum height=0.9cm, minimum width=2.2cm, inner sep=4pt, align=center,
    font=\sffamily\small, text=cyggrey},
  passarr/.style={->, draw=cyggreen, line width=1.1pt},
  failarr/.style={->, draw=cygred, line width=1pt},
  skiparr/.style={->, draw=cyggrey!50, line width=0.8pt, dashed},
  dataarr/.style={->, draw=black!65, line width=0.9pt},
  elab/.style={font=\sffamily\scriptsize, inner sep=1.5pt, fill=white, fill opacity=0.85, text opacity=1},
}

\fill[cyglblue, rounded corners=5pt] (0, 5.8) rectangle (22, 10.2);
\node[anchor=north west, text=cygblue, font=\sffamily\bfseries\small] at (0.2, 10.1) {Short-circuit mode};

\node[font=\sffamily\small] (q1) at (-1.2, 8.0) {Query};

\node[vbox] (sc-regex) at (1.8, 8.0) {Regex};
\node[vbox] (sc-ast) at (5.6, 8.0) {AST};
\node[resultbox] (sc-explain) at (9.6, 8.0) {EXPLAIN};
\node[skipbox] (sc-mirror) at (14.0, 8.0) {Mirror-Execute};

\node[resultbox] (sc-result) at (19.0, 8.0) {Schema error};

\draw[dataarr] (q1.east) -- (sc-regex.west);
\draw[passarr] (sc-regex.east) -- node[above, elab, text=cyggreen] {pass} (sc-ast.west);
\draw[passarr] (sc-ast.east) -- node[above, elab, text=cyggreen] {pass} (sc-explain.west);
\draw[skiparr] (sc-explain.east) -- node[above, elab, text=cyggrey] {skipped} (sc-mirror.west);
\draw[skiparr] (sc-mirror.east) -- node[above, elab, text=cyggrey] {skipped} (sc-result.west);

\node[font=\sffamily\scriptsize, text=cygred] at (9.6, 6.8) {chain stops};
\draw[->, draw=cygred, line width=0.8pt] (sc-explain.south) -- (9.6, 7.05);

\draw[failarr] (sc-explain.south) .. controls (9.6, 6.4) and (19.0, 6.4) .. (sc-result.south);

\fill[cyglblue, rounded corners=5pt] (0, 0.2) rectangle (22, 4.8);
\node[anchor=north west, text=cygblue, font=\sffamily\bfseries\small] at (0.2, 4.7) {Collect-all mode};

\node[font=\sffamily\small] (q2) at (-1.2, 2.5) {Query};

\node[vbox] (ca-regex) at (1.8, 2.5) {Regex};
\node[passbox] (ca-ast) at (5.6, 2.5) {AST};
\node[resultbox] (ca-explain) at (9.6, 2.5) {EXPLAIN};
\node[resultbox] (ca-mirror) at (14.0, 2.5) {Mirror-Execute};

\node[resultbox, minimum width=5cm] (ca-result) at (19.2, 2.5) {Ordered error list};

\draw[dataarr] (q2.east) -- (ca-regex.west);
\draw[passarr] (ca-regex.east) -- node[above, elab, text=cyggreen] {pass} (ca-ast.west);
\draw[passarr] (ca-ast.east) -- node[above, elab, text=cyggreen] {pass} (ca-explain.west);
\draw[dataarr] (ca-explain.east) -- (ca-mirror.west);
\draw[dataarr] (ca-mirror.east) -- (ca-result.west);

\draw[failarr, dashed] (ca-explain.south) .. controls (9.6, 1.2) and (17.0, 1.2) .. node[below, midway, elab, text=cygred] {schema error} (ca-result.south west);
\draw[failarr, dashed] (ca-mirror.south) .. controls (14.0, 1.4) and (18.0, 1.4) .. node[below, near start, elab, text=cygred] {property error} (ca-result.south);

\end{tikzpicture}
\caption{Validator chain modes. In short-circuit mode (top) the chain stops at the first failure and returns a single error. In collect-all mode (bottom) the chain runs every backend whose precondition is satisfied and aggregates all findings into an ordered list.}
\label{fig:chain-semantics}
\end{figure}

The four backends compose into a validator chain that operates in one of two modes. In short-circuit mode (default), the chain runs backends in order and returns the first failure. The default order is regex, AST, EXPLAIN, mirror-execute, placing the cheapest filter first and the most authoritative check last. In collect-all mode, the chain runs every backend whose precondition is satisfied and aggregates the results. Backends are skipped when an earlier backend has already reported a parse error for the same query. When the AST backend reports a parse failure but the EXPLAIN backend successfully plans the same query, the chain treats the AST result as a grammar-coverage gap and promotes the EXPLAIN outcome.

The result list in collect-all mode is ordered by a fixed contract. Parse errors come first because a parse failure dominates other findings. Within categories, errors are ordered by the authority of the backend that produced them, with EXPLAIN and mirror-execute above AST and regex. Within the same backend, errors are ordered by severity, with constraint violations above property-type errors above schema references.

\subsection{Mirror graph and equivalence verification}
\label{sec:methods-mirror}

We adapt the technique of using a synthetic graph as a Cypher validation environment \citep{tiwari2024synthcypher,zhong2025synthet2c} to a runtime feedback role. The mirror graph is a separate Neo4j instance whose schema matches the target graph but whose data is minimal. The mirror is built once from the schema and reused across many query validations, and the per-query overhead becomes the cost of a planning round-trip rather than the cost of data generation. The gate constructs one node per declared label with type-correct dummy values populated on each property, and one relationship per declared relationship type connecting one source-label node to one target-label node. Constraints and indexes are applied as data definition language (DDL) statements\footnote{Some constraint types such as existence and node-key constraints are enterprise-only features in Neo4j and are skipped when the mirror runs on Community Edition.}. The mirror serves as the EXPLAIN target so the structural validator can plan queries without touching production, and as the execution target for the mirror-execute backend so runtime errors are caught without affecting production.

A mirror constructed from a YAML schema specification must produce validation outcomes equivalent to those produced against the source database. We verify this by running an equivalence procedure against ten reference databases. The procedure compares validation outcomes across three targets for the same query sample, a YAML-built mirror, an introspection-built mirror, and the source database. Outcomes are compared on verdict (passed or failed), failure category (parse, schema, property, or constraint), and the unknown reference reported on schema-category failures. Plan shapes, row estimates, and cost estimates are not compared as they are expected to differ between a minimal mirror and a populated source database. The equivalence claim is about structural validation, not about cost.

The query sample contains seventeen templates covering common Cypher patterns. Nine templates are success-shape (single-node match, path patterns with one or more relationships, variable-length bounded paths, aggregation with count and group-by, ORDER BY with LIMIT, WHERE with multiple predicates, OPTIONAL MATCH). Five are failure-shape (unknown label, unknown relationship type, unknown property, property type mismatch, and two parse failures testing missing parenthesis and missing keyword). Three are edge cases (self-loop relationship, pattern comprehension, CASE WHEN expression). The templates are parameterised by the target schema so the same sample exercises any database with one label, one relationship type, and one property of each common type. The ten reference databases span schema-shape variety including multi-label nodes, sparse-optional properties, self-loops, hub-shaped cardinalities, composite constraints, datetime properties, hierarchical taxonomies, and recursive relationships.

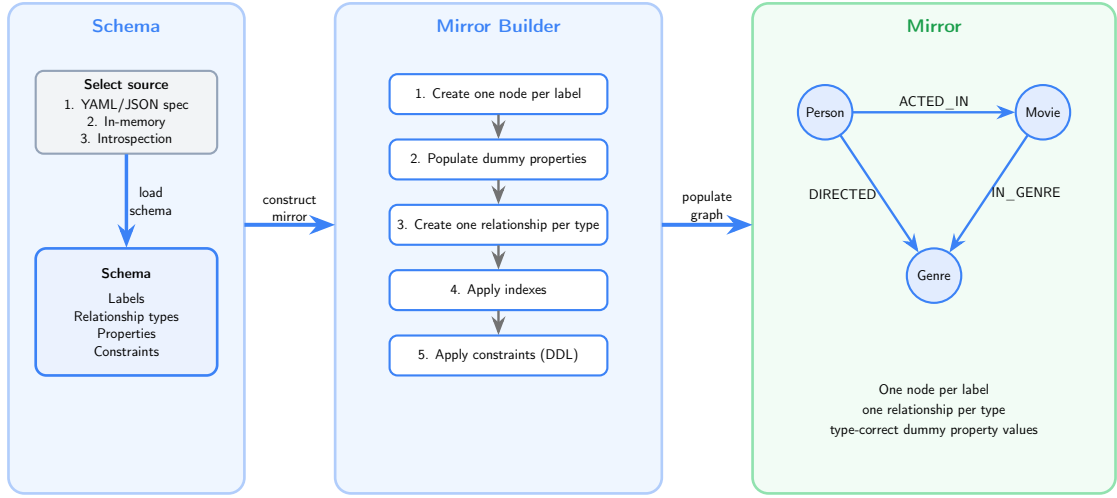
\begin{figure}[h]
\centering
\begin{tikzpicture}[>=Stealth, font=\sffamily\normalsize, scale=0.48, transform shape]
 
\definecolor{cygblue}{HTML}{3B82F6}
\definecolor{cyggreen}{HTML}{22C55E}
\definecolor{cyggrey}{HTML}{94A3B8}
\definecolor{cyglblue}{HTML}{EFF6FF}
 
\tikzset{
  srcbox/.style={rounded corners=3pt, draw=cyggrey, fill=cyggrey!12, line width=0.8pt,
    minimum height=1.6cm, minimum width=5cm, inner sep=6pt, align=center,
    font=\sffamily\normalsize},
  schemabox/.style={rounded corners=4pt, draw=cygblue, fill=cygblue!10, line width=1pt,
    minimum height=3.5cm, minimum width=5cm, inner sep=8pt, align=center,
    font=\sffamily\normalsize},
  stepbox/.style={rounded corners=3pt, draw=cygblue, fill=white, line width=0.8pt,
    minimum height=1.1cm, minimum width=6cm, inner sep=5pt, align=center,
    font=\sffamily\normalsize},
  gnode/.style={circle, draw=cygblue, fill=cygblue!15, line width=0.8pt,
    minimum size=1.5cm, inner sep=2pt, font=\sffamily\normalsize, align=center},
  section/.style={rounded corners=6pt, line width=1pt},
  bluearr/.style={->, draw=cygblue, line width=1.3pt},
  dataarr/.style={->, draw=black!55, line width=0.9pt},
  elab/.style={font=\sffamily\normalsize, inner sep=2pt},
}
 
\fill[cyglblue, rounded corners=6pt] (-0.5, -0.5) rectangle (6, 13);
\draw[section, draw=cygblue!40] (-0.5, -0.5) rectangle (6, 13);
\node[font=\sffamily\bfseries\Large, text=cygblue] at (2.75, 12.4) {Schema};
 
\fill[cyglblue, rounded corners=6pt] (8.5, -0.5) rectangle (17.5, 13);
\draw[section, draw=cygblue!40] (8.5, -0.5) rectangle (17.5, 13);
\node[font=\sffamily\bfseries\Large, text=cygblue] at (13.0, 12.4) {Mirror Builder};
 
\fill[cyggreen!6, rounded corners=6pt] (20, -0.5) rectangle (30, 13);
\draw[section, draw=cyggreen!50] (20, -0.5) rectangle (30, 13);
\node[font=\sffamily\bfseries\Large, text=cyggreen!80!black] at (25.0, 12.4) {Mirror};
 
\node[srcbox] (select) at (2.75, 10.0) {\textbf{Select source}\\[2pt]{\normalsize 1. YAML/JSON spec}\\{\normalsize 2. In-memory}\\{\normalsize 3. Introspection}};
 
\node[schemabox] (schema) at (2.75, 4.5) {\textbf{Schema}\\[6pt]{\normalsize Labels}\\{\normalsize Relationship types}\\{\normalsize Properties}\\{\normalsize Constraints}};
 
\draw[bluearr] (select.south) -- node[right, elab, align=center] {load\\schema} (schema.north);
 
\node[stepbox] (s1) at (13.0, 10.5) {1. Create one node per label};
\node[stepbox] (s2) at (13.0, 8.7)  {2. Populate dummy properties};
\node[stepbox] (s3) at (13.0, 6.9)  {3. Create one relationship per type};
\node[stepbox] (s4) at (13.0, 5.1)  {4. Apply indexes};
\node[stepbox] (s5) at (13.0, 3.3)  {5. Apply constraints (DDL)};
 
\draw[dataarr] (s1.south) -- (s2.north);
\draw[dataarr] (s2.south) -- (s3.north);
\draw[dataarr] (s3.south) -- (s4.north);
\draw[dataarr] (s4.south) -- (s5.north);
 
\node[gnode] (person) at (22.0, 10.0) {Person};
\node[gnode] (movie)  at (28.0, 10.0) {Movie};
\node[gnode] (genre)  at (25.0, 5.5)  {Genre};
 
\draw[->, draw=cygblue, line width=0.9pt] (person) -- node[above, elab] {ACTED\_IN} (movie);
\draw[->, draw=cygblue, line width=0.9pt] (movie) -- node[right, elab] {IN\_GENRE} (genre);
\draw[->, draw=cygblue, line width=0.9pt] (person) -- node[left, elab] {DIRECTED} (genre);
 
\node[font=\sffamily\normalsize, text=black, align=center] at (25.0, 1.8)
  {One node per label\\[2pt]one relationship per type\\[2pt]type-correct dummy property values};
 
\draw[bluearr, line width=1.5pt] (6, 6.9) -- node[above, elab, align=center] {construct\\mirror} (8.5, 6.9);
 
\draw[bluearr, line width=1.5pt] (17.5, 6.9) -- node[above, elab, align=center] {populate\\graph} (20, 6.9);
 
\end{tikzpicture}
\caption{Mirror graph construction from a schema description. A schema loaded from one of three sources drives a five-step construction process that populates a minimal Neo4j instance with one node per declared label and one relationship per declared type.}
\label{fig:mirror-construction}
\end{figure}

\subsection{Cost gate}
\label{sec:methods-cost-gate}

The cost gate uses Neo4j's EXPLAIN planner output to estimate the expense of a query before it executes. The planner returns a plan tree where each node carries an estimated row count and an estimated database-hit count. The cost gate parses this tree, identifies the operators with the highest row estimates, and compares the overall plan estimate against a configured threshold. Queries exceeding the threshold are rejected with a structured error that names the cost driver operator, the variables it operates on, and a list of suggested mitigations specific to that operator. Operators that motivate specific mitigations include all-nodes-scans (suggest a label filter), cartesian products (suggest a relationship), variable-length paths with high estimates (suggest a depth bound), and unbounded expansions (suggest a relationship type filter or a result limit).

The cost-driver heuristic operates on a priority-then-depth ordering rather than pure deepest-wins. Some operators (cartesian product, variable-length expand) are actionable for the agent that generated the query, while others (the planner's terminal projection nodes) are not. The priority list places actionable operators first, with depth breaking ties within each priority bucket. Without this ordering the heuristic returns operators that name a non-actionable cost driver more often than it returns operators an agent can fix.

The cost gate emits a structured payload that carries the estimated rows, the estimated database-hits, the threshold used, the cost driver, and a top-N breakdown of operators with their individual estimates. The breakdown lets the downstream consumer reason about cost beyond a single driver.

The methodology of pre-execution cost estimation for an agent's generated queries is established in adjacent literature. Recent work on cost-aware text-to-SQL \citep{deochake2025costaware} measures bytes-processed, slot-utilisation, and dollar cost per query on BigQuery and reports that LLM-generated SQL exhibits cost variance of more than three times across models. The pattern of pre-execution cost estimation for graph queries appears in research on GraphQL \citep{mavroudeas2021graphql} where machine learning is applied to estimate query expense. We apply the methodology to Cypher with the planner's own estimates as the cost signal, using a static threshold rather than a learned estimator.

\subsection{Error vocabulary}
\label{sec:methods-error-vocab}

The gate classifies every failure into one of five categories. Parse errors describe syntactic problems and carry line and column information together with an excerpt of the failing query region so the corrector can locate the problem. Schema errors describe references to labels, relationships, or properties that do not exist in the schema and carry did-you-mean suggestions based on string similarity together with a list of the valid schema elements at that point in the query. Property errors describe type mismatches where a literal of one type is compared against a property declared as a different type, and they carry the declared and observed types together with did-you-mean suggestions for the property name. Constraint errors describe violations of uniqueness or existence constraints on CREATE and MERGE patterns. Cost errors describe rejections by the cost gate as described above. Empty-result errors are defined in the vocabulary and reserved for the optional empty-result detector.

The structured error vocabulary is the contract between the gate and any downstream refinement system. The same vocabulary is consumed by the gate's own corrector and is available to external correctors through all three transports. The five-category taxonomy is not present as a controlled vocabulary in prior Cypher tooling. CyVer \citep{mandilara2025cyver} surfaces errors with metadata but does not specify a fixed category vocabulary or a structured payload across categories. The vocabulary is one of this work's design contributions, drawn from the SQL self-correction literature \citep{wang2023macsql,ni2023lever} where structured execution feedback is recognised as a substantive component of refinement quality.

\subsection{Corrector}
\label{sec:methods-corrector}

The corrector follows the pattern from MAC-SQL \citep{wang2023macsql} adapted to Cypher and to the structured error vocabulary above. It accepts a failed query, the gate's error description, and a context object containing the schema and any prior refinement attempts, and returns either a refined query that the agent should retry or an abort signal. The corrector interface is a single-shot protocol so callers can plug in their own implementation. Each corrector call produces one refinement attempt. We implement five LLM-backed correctors and a null corrector that always returns abort for use as a safe default when no refinement is configured.

The five correctors differ in how they assemble the prompt for the language model. The Raw corrector passes the Neo4j error string as-is alongside a flat schema summary listing all declared labels, relationship types, and properties. The Verbal corrector renders the gate's structured error payload as a natural-language prose paragraph alongside the same flat summary. The FullSchema corrector passes the entire schema to the model without any filtering or relevance gating. The BudgetSchema corrector passes a truncated schema at a fixed token budget, simulating the trade-off a caller faces on limited-context models where the full schema would exceed the available window. The RAMPART corrector assembles its prompt from stored prompt blocks using the RAMPART block registry \citep{tomczak2026rampart} with per-call filtering and token-budgeted compilation, described below.

The block-based corrector rebuilds a fresh registry on every call. The block-selection mechanism is rule-based rather than relevance-scored. The error-vocabulary block is selected by exact string match on the failure's error category, so a schema-category failure loads the schema error block and a parse-category failure loads the parse error block. Only the matching category's block is loaded. Schema blocks are filtered to the entities the failing query references through a regex-based entity extractor that identifies label tokens in the query text, classifies each against declared schema labels and relationship types, folds the error payload's did-you-mean and available-in-scope lists into the relevant set, and includes every relationship type whose source or target label is already in the relevant set (a one-hop expansion). A query that resolves to zero specific references falls back to the full schema. Prior refinement attempts are written as separate evictable blocks carrying both the refined query and its full structured error payload. The blocks are compiled to a configured token budget with eviction of lower-priority tail blocks when the budget is exceeded.

The block-based assembly uses three capabilities from the registry, namely the typed container, position-ordered compilation, and token-budget eviction. The relevance filtering (regex entity extraction, exact-match category selection, one-hop schema expansion) is corrector-specific logic rather than a registry feature.

The corrector enforces a strict response contract on the language model. Responses that fail the contract are handled by a protocol-retry layer that re-asks the model with an escalated reminder of the response shape, up to a configurable retry limit, before aborting. The protocol-retry layer wraps the corrector as a composable decorator rather than living inside any specific corrector, so all five correctors benefit from it. The separation of protocol compliance from refinement quality lets the evaluation distinguish cases where the model returned a malformed response from cases where the model returned correctly shaped Cypher that does not fix the error, which are different failure modes with different mitigations.

A refinement loop iterates the corrector up to a configurable number of attempts (default three), stopping when the validator chain passes or the attempt budget is exhausted. On each iteration the loop re-validates the corrector's output against the validator chain. If validation fails, the loop constructs a new error context from the fresh validation failure and feeds the failed attempt back to the corrector as a prior-attempt block. For the block-based corrector this means the error-vocabulary block may change between attempts (if the first attempt fixes a schema reference but introduces a property mismatch, the second attempt loads the property-error vocabulary instead of the schema-error vocabulary), and the schema-relevance filtering is recomputed against the revised query and the new error.

Three retry layers operate at distinct levels. Transport retry in the resilient LLM client handles HTTP errors (429 rate limits with long backoff, 503 transient failures with exponential backoff, timeouts). Protocol retry in the decorator handles malformed LLM responses (wrong JSON shape, truncated output). Refinement retry in the loop handles valid-but-wrong Cypher. The three layers are separate and composable.

\begin{table}[h]
\centering
\caption{Prompt structure of the five correctors. All share the same system prompt (provided in the appendix). They differ in schema presentation, error rendering, and whether prior attempts and token-budget eviction are applied.}
\label{tab:corrector-prompts}
\small
\tightcols
\begin{tabularx}{\textwidth}{@{}p{2.4cm} >{\raggedright\arraybackslash}X >{\raggedright\arraybackslash}X >{\raggedright\arraybackslash}X >{\raggedright\arraybackslash}X >{\raggedright\arraybackslash}X@{}}
\toprule
Component & Raw & Verbal & FullSchema & BudgetSchema & RAMPART \\
\midrule
Schema & Flat summary (labels + rel types) & Flat summary (same as Raw) & Complete (all labels, all properties, all rel types) & Truncated to token budget & Per-entity blocks, relevance-filtered to query \\
\addlinespace
Error & Raw Neo4j error string & Prose paragraph from structured payload & Structured JSON payload & Structured JSON payload & Category-matched error-vocabulary block \\
\addlinespace
Prior attempts & No & No & No & No & Yes (failed attempts as blocks) \\
\addlinespace
Token eviction & No & No & No & Yes (fixed budget) & Yes (tail blocks evicted) \\
\addlinespace
Relevance filter & No & No & No & No & Yes (schema filtered to query entities) \\
\bottomrule
\end{tabularx}
\end{table}

We evaluate corrector performance through a two-component decomposition that separates protocol compliance from refinement quality. The prompt-following rate is the fraction of queries on which the model produced any evaluable Cypher (as opposed to returning malformed JSON or non-Cypher text). The Cypher-repair rate is the fraction of evaluable responses whose refined Cypher passes the validator chain. The overall success rate is the product of the two. This decomposition matters because the two failure modes have different causes and different mitigations. A model that produces correct Cypher 95\% of the time but follows the response protocol only 50\% of the time achieves 47.5\% overall, ranking below a weaker model with 70\% Cypher quality and 90\% prompt-following at 63\% overall. Reporting overall success alone would attribute the protocol deficit to the model's Cypher capability.

The system is available as an installable Python package at \url{https://github.com/softmatsg/thulge-cygnet-rel}.

\section{Results and discussion}
\label{sec:results}

\subsection{Mirror equivalence and error vocabulary}
\label{sec:results-functioning}

As a control for the experiments that follow, we verified mirror equivalence across ten reference databases of varying schema shapes per the methodology in Section~\ref{sec:methods-mirror}. Seventeen of seventeen template queries produce identical validation outcomes against YAML-built mirror, introspection-built mirror, and source database on every database tested. Three schema-shape asymmetries were observed and documented as known limitations. Indexes that back uniqueness constraints are not modelled as separate schema elements because they are inherently tied to their owning constraint. Composite node-key constraints across multiple properties are not modelled by the current schema representation. Spatial Point types are not modelled and are represented as paired float coordinates in the test schemas. None of the three affects validation outcomes on the templates tested.

Example~\ref{lst:schema-err} shows a schema-error query and the corrected version produced by an agent given the gate's structured error. The unknown label \texttt{Smaple} is identified and a did-you-mean suggestion is returned in the structured payload. The corrected query passes validation.

\begin{lstlisting}[caption=Schema reference error with did-you-mean suggestion.,label=lst:schema-err]
<@\textsf{\textbf{1. Faulty query}}@>
   MATCH (s:Smaple) RETURN s.id

<@\textsf{\textbf{2. Gate error}}@>
   SchemaError(
     reference_kind="label",
     unknown_reference="Smaple",
     did_you_mean=["Sample"],
     available_in_scope=["Sample","Measurement","Compound"])

<@\textsf{\textbf{3. Refined query (passes validation)}}@>
   MATCH (s:Sample) RETURN s.id
\end{lstlisting}

Example~\ref{lst:cost-err} shows a cost-rejected query and the mitigation the gate suggests. The cartesian product across two disconnected patterns produces an estimated row count exceeding the configured threshold.

\begin{lstlisting}[caption=Cost rejection with operator-specific mitigation.,label=lst:cost-err]
<@\textsf{\textbf{1. Faulty query}}@>
   MATCH (a:Actor), (m:Movie) RETURN a, m

<@\textsf{\textbf{2. Gate error}}@>
   CostError(
     estimated_rows=2400000,
     estimated_dbhits=4800000,
     threshold_used=100000,
     cost_driver="CartesianProduct",
     suggested_mitigations=[
       "Add a relationship between disconnected variables."])

<@\textsf{\textbf{3. Refined query (passes validation)}}@>
   MATCH (a:Actor)-[:ACTED_IN]->(m:Movie) RETURN a, m
\end{lstlisting}

\subsection{Validator quality}
\label{sec:results-validator-quality}

Existing Cypher corpora label execution correctness or syntax errors but do not distinguish parse, schema-reference, property-type, and constraint failures. We generated template-based test corpora for nine schemas of varying shape. For each schema a corpus was generated programmatically with known ground-truth labels. Valid queries were constructed from the schema's declared labels, properties, and relationship types. Broken queries were produced by injecting deliberate errors in four categories. Parse errors introduced syntax breaks (missing keywords, unbalanced parentheses). Schema errors replaced a label or relationship type with a valid sibling from the same schema, simulating the plausible-but-wrong references that real LLMs produce rather than obvious garbage. Path-form schema errors swap a label or relationship type in a query that traverses a relationship pattern with labelled endpoints on both sides. Standalone schema errors swap a label in a query with no relationship. Property errors replaced a property name with a different valid property from the same label. Constraint errors violated declared uniqueness or existence constraints. The total corpus across nine schemas is 1135 queries.

\begin{table}[h]
\centering
\caption{Validator true-positive rate (TPR) across nine schemas with template-generated corpora. The false-positive rate is zero on every schema and every backend (no valid query was ever flagged as broken).}
\label{tab:cross-schema-tpr}
\small
\begin{tabular}{lrrrrrr}
\toprule
Schema & builtin & AST & EXPLAIN & mirror-exec & chain & corpus \\
\midrule
movies\_recommendations  & 0.095 & 0.571 & 0.262 & 0.286 & \textbf{0.643} & 140 \\
northwind                & 0.170 & 0.723 & 0.234 & 0.362 & \textbf{0.766} & 203 \\
stackoverflow            & 0.163 & 0.651 & 0.233 & 0.279 & \textbf{0.698} & 136 \\
twitter                  & 0.200 & 0.675 & 0.275 & 0.300 & \textbf{0.725} & 100 \\
pole\_crime              & 0.163 & 0.674 & 0.233 & 0.279 & \textbf{0.698} & 137 \\
graphacademy\_fraud      & 0.175 & 0.675 & 0.250 & 0.275 & \textbf{0.675} & 115 \\
mekg                     & 0.156 & 0.600 & 0.222 & 0.311 & \textbf{0.711} & 171 \\
mesh\_ontology           & 0.206 & 0.618 & 0.294 & 0.265 & \textbf{0.618} & 71 \\
multi\_label\_synthetic  & 0.138 & 0.414 & 0.345 & 0.276 & \textbf{0.448} & 62 \\
\midrule
mean                     & 0.163 & 0.622 & 0.261 & 0.292 & \textbf{0.665} & 1135 \\
\bottomrule
\end{tabular}
\end{table}

The false-positive rate (FPR) is zero across all nine schemas and all conditions (Table~\ref{tab:cross-schema-tpr}). No valid query was ever flagged as broken. On an external corpus of 683 independently authored valid queries \citep{bratanic2024synthetic} the chain flagged 39. All 39 are valid Cypher 5+ syntax that the bundled Cypher 9 grammar does not parse, most prominently count and exists subquery expressions. No structurally valid query was incorrectly rejected.

\begin{table}[h]
\centering
\caption{Per-error-category recall for the chain across nine schemas. Detection depends on whether the injected error produces structurally distinguishable Cypher.}
\label{tab:per-category-recall}
\small
\begin{tabularx}{\textwidth}{l r r X}
\toprule
Category and sub-class & n & Recall & Detection mechanism \\
\midrule
parse (syntax breaks)                          & 63 & \textbf{1.00} & AST parser catches missing keywords and unbalanced constructs \\
constraint (uniqueness and existence)           & 41 & \textbf{1.00} & mirror-execute catches violations \\
\addlinespace
schema, garbage label/rel (non-existent name)   & 18 & \textbf{1.00} & reference absent from schema \\
schema, path-form swap (labelled endpoints)     & 88 & \textbf{1.00} & rel-endpoint compatibility check verifies the relationship connects the bound labels \\
schema, standalone swap (no relationship)       & 48 & 0.48 & cascading property check catches swaps where the query accesses a property absent from the new label \\
schema, path-form swap (unlabelled endpoint)    & 42 & 0.00 & endpoint check skips when a node has no label \\
\addlinespace
property, sibling-swap (valid on target label)  & 63 & 0.00 & structurally undetectable (both properties exist on the label) \\
\bottomrule
\end{tabularx}
\end{table}

Parse errors and constraint violations are detected at 100\% and 100\% respectively (Table~\ref{tab:per-category-recall}). The chain catches 100\% of schema errors where the injected name does not exist in the schema and 100\% of path-form swaps where both endpoints carry labels, through a rel-endpoint compatibility check that verifies the relationship type connects the bound labels in the schema. For standalone node queries where a label is swapped but no relationship is traversed, the chain catches 48\% through a cascading mechanism where the query accesses a property of the original label that does not exist on the substituted label. When no property is accessed, the swap is structurally invisible. Path-form swaps where one endpoint has no label score 0\% because the endpoint check cannot determine the intended label. Property sibling-swaps where the substituted property also exists on the target label are structurally undetectable by definition.

The AST backend is the strongest individual backend. The chain TPR of 0.665 reflects the union of all backends. The mirror-execute backend is essential for constraint-category recall (1.00 vs 0.65 for AST alone) but contributes little incremental lift on other categories.

\subsection{EXPLAIN calibration}
\label{sec:results-calibration}

The cost gate uses Neo4j's EXPLAIN command, which asks the query planner to produce an execution plan with estimated row counts without actually running the query. The gate compares the planner's estimated rows against a configurable threshold and rejects queries whose estimates exceed it. The credibility of this approach rests on how accurately the planner's estimates track actual row counts. We measured this on the recommendations dataset over a programmatically generated query corpus of 300 queries spanning seven query patterns of increasing structural complexity.

\begin{table}[h]
\centering
\caption{Planner EXPLAIN calibration against actual row counts on the recommendations dataset (28863 nodes). Ratio is estimated rows divided by actual rows (1.0 means the planner estimated exactly the number of rows returned). Within-$k$x is the fraction of queries whose ratio falls in $[1/k, k]$. Cartesian products report 9 of 20 completed (11 timed out at the 30-second per-query limit).}
\label{tab:calibration}
\small
\begin{tabular}{lrrrrrr}
\toprule
Query pattern & n & Median ratio & p25 & p75 & Within 2x (\%) & Within 10x (\%) \\
\midrule
match-return              & 50  & 1.0 & 1.0 & 1.0 & 100.0 & 100.0 \\
aggregation               & 50  & 1.0 & 1.0 & 1.0 & 100.0 & 100.0 \\
single-hop                & 50  & 1.0 & 1.0 & 3.0 & 68.0 & 100.0 \\
multi-hop bounded         & 50  & 3.0 & 2.2 & 3.0 & 24.0 & 94.0 \\
match-where-return        & 50  & 39 & 0.4 & 57 & 28.0 & 32.0 \\
var-length unbounded      & 30  & $>10^{6}$ & 3.2 & $>10^{6}$ & 0.0 & 33.3 \\
Cartesian (completed)     & 9   & 1.0 & 1.0 & 1.0 & 100.0 & 100.0 \\
\midrule
overall                   & 300 & 1.0 & & & 56.3 & 77.3 \\
\bottomrule
\end{tabular}
\end{table}

The overall median ratio is 1.0, meaning the planner's row estimate matches the actual row count exactly for the typical query (Table~\ref{tab:calibration}). The planner calibrates well on four of seven patterns, with match-return, aggregation, single-hop, and Cartesian all achieving median ratios of 1.0. Multi-hop bounded queries show moderate overestimation (median 3.0, 94\% within 10x). Two patterns are explicit outliers. Match-where-return has a median ratio of 39, reflecting the planner's difficulty estimating selectivity on property predicates against the dataset's skewed value distributions. Variable-length unbounded queries overestimate by orders of magnitude because the planner cannot bound the cardinality of uncapped traversals. The cost gate's value on these outlier patterns is in flagging them as expensive before execution, which is the conservative-correct behaviour for a pre-execution gate, rather than in providing accurate cost estimates. Eleven of twenty Cartesian product queries timed out at the 30-second per-query execution limit set for the benchmark. The nine that completed all had ratio 1.0, confirming the planner correctly estimates Cartesian cardinality when execution completes.

\subsection{Latency}
\label{sec:results-latency}

We measured per-stage gate latency on a 500-query corpus drawn from the same generator as the calibration study, with each query timed five times, against the recommendations dataset.

\begin{table}[h]
\centering
\caption{Per-stage and full-pipeline gate latency on a 500-query corpus against the recommendations dataset. Each query timed five times (2500 measurements total).}
\label{tab:latency}
\begin{tabular}{lrrr}
\toprule
Stage & p50 (ms) & p95 (ms) & p99 (ms) \\
\midrule
parse (regex)           & 0.052 & 0.129 & 0.198 \\
ANTLR parse             & 2.077 & 4.156 & 5.808 \\
AST schema check        & $<$0.001 & 0.956 & 2.026 \\
EXPLAIN round-trip      & 2.214 & 4.909 & 23.037 \\
full pipeline           & 5.576 & 9.243 & 11.954 \\
failed-query fast path  & 0.141 & 1.293 & 2.837 \\
\bottomrule
\end{tabular}
\end{table}

The median full pipeline takes 5.6 ms (Table~\ref{tab:latency}). The 99th-percentile latency of 12.0 ms leaves headroom for use in interactive agent loops where typical LLM inference takes one to several seconds. The ANTLR parser and the EXPLAIN round-trip each contribute approximately 2.1-2.2 ms at the median, together accounting for most of the pipeline time. Failed queries complete at a median of 0.14 ms, approximately 40 times faster than successful queries, reflecting early termination of the chain on the first definitive failure.

\subsection{Multi-error collection mode}
\label{sec:results-multi-error}

We measured the effect of the collect-all chain mode on a full-corpus run against the movies subset of the corpus. A broken query may contain one or more distinct structural problems (for example a misspelled label and a missing keyword). Each backend that detects a problem produces an error report. A single distinct problem may therefore generate multiple reports when several backends detect it independently with different wordings. The short-circuit mode stops at the first backend that reports an error and returns that single report. The collect-all mode runs every backend whose precondition is satisfied and aggregates all reports. The two modes catch the same set of broken queries since the underlying backends are identical. The difference is in how many reports the caller receives per broken query.

In short-circuit mode each broken query produces exactly one report by construction. In collect-all mode the chain surfaces a mean of 2.57 reports per broken query, corresponding to 1.11 distinct problems and 2.31 reports per distinct problem. Eighty-nine percent of broken queries contain exactly one distinct problem that multiple backends identify with different wordings. The remaining 11.0\% contain two or more distinct problems.

For the 11.0\% of queries with multiple distinct problems, the collect-all mode halves the agent loop count an automatic corrector requires to fully fix the query, because the corrector sees both problems in a single round-trip. For the remaining 89.0\%, the collect-all mode produces multiple phrasings of the same underlying problem, varying by backend because each backend identifies and presents the error differently. The built-in regex backend reports the unknown reference and a did-you-mean suggestion. The AST backend reports the same reference with line and column information. The EXPLAIN backend reports it from the Neo4j planner's perspective with the available references in scope at the failure site. The mirror-execute backend reports it as a runtime semantic error. Whether the diversity of phrasings improves refinement quality is a downstream question that depends on the specific language model consuming the report. Both modes are available and the choice is left to the caller.

\begin{table}[h]
\centering
\caption{Chain mode comparison on the nine-schema corpus. Both modes catch the same set of broken queries. The difference is in the per-query error payload and latency.}
\label{tab:multi-error}
\begin{tabular}{lrrrr}
\toprule
Chain mode & Mean reports & Mean distinct & Single-problem (\%) & p50 (ms) \\
\midrule
short-circuit & 1.00 & 1.00 & n/a & 0.202 \\
collect-all   & 2.57 & 1.11 & 89.0 & 11.217 \\
\bottomrule
\end{tabular}
\end{table}

The latency cost of collect-all mode is substantial in relative terms but small in absolute terms. Median full-pipeline latency rises from 0.2 ms in short-circuit mode to 11.2 ms in collect-all mode (Table~\ref{tab:multi-error}). The short-circuit median is low because 76\% of queries are flagged at the built-in stage and never reach EXPLAIN. Both modes remain well under the 50 ms target.

The default chain configuration is short-circuit. The collect-all mode is available as a per-call option. Deployments where refinement quality matters more than per-call latency, particularly correctors driven by small or quantised local models that benefit from prompt diversity, have evidence to prefer the collect-all mode. High-throughput pipelines against a known schema have evidence to prefer the short-circuit default.

\subsection{Corrector performance}
\label{sec:results-corrector}

Table~\ref{tab:corrector-comparison} reports overall success rates for all five correctors on a stratified corpus of 98 broken queries (24 parse errors, 74 schema errors) across five Google-family models, with up to 3 refinement attempts per query.

\begin{table}[h]
\centering
\caption{Overall success rate (\%) per corrector and model on the stratified corpus ($n = 98$), after up to 3 refinement attempts.}
\label{tab:corrector-comparison}
\small
\begin{tabular}{lrrrrr}
\toprule
Model & RAMPART & Raw & Verbal & FullSchema & BudgetSchema \\
\midrule
Gemini 3.1 Pro        & 94.9 & 92.9 & \textbf{98.0} & 94.9 & 93.9 \\
Gemini 3 Flash        & \textbf{93.9} & 85.7 & 92.9 & 90.8 & 87.8 \\
Gemini 2.5 Flash Lite & 80.6 & 74.5 & 90.8 & \textbf{92.9} & 91.8 \\
Gemma 4 26B           & \textbf{92.9} & 77.6 & 80.6 & 85.7 & 84.7 \\
Gemma 4 31B           & \textbf{82.7} & 44.9 & 75.5 & 49.0 & 50.0 \\
\midrule
Mean                  & \textbf{89.0} & 75.1 & 87.6 & 82.7 & 81.6 \\
\bottomrule
\end{tabular}
\end{table}

RAMPART achieves the highest mean at 89.0\% but Verbal is within 1.4 percentage points at 87.6\% with approximately one fifth of RAMPART's input-token cost. On individual models, Verbal leads on Pro (98.0\%) and Flash Lite (90.8\%), while RAMPART leads on Flash (93.9\%), Gemma 26B (92.9\%), and Gemma 31B (82.7\%). The FullSchema corrector reaches 92.9\% on Flash Lite and ties RAMPART at 94.9\% on Pro, confirming that on large-context models the full schema performs as well as relevance-filtered assembly. This is consistent with the finding that schema filtering benefits smaller models more than larger ones \citep{ozsoy2025schemafilter}. The BudgetSchema corrector tracks FullSchema within 1-3 percentage points on every model, indicating that on this schema the token budget did not bind.

Gemma 4 31B is the most sensitive to corrector choice, with a 37.8 percentage-point gap between RAMPART (82.7\%) and Raw (44.9\%). Without structured-output enforcement Gemma 31B produces malformed responses at rates between 79\% and 89\% per call. No single corrector dominates across all models. The per-corrector gap is driven primarily by prompt-following rather than Cypher-repair quality.

Table~\ref{tab:per-category} reports per-category success rates for the RAMPART corrector.

\begin{table}[h]
\centering
\caption{Per-category success rates for the RAMPART corrector on the stratified corpus.}
\label{tab:per-category}
\begin{tabular}{lrr}
\toprule
Model & Parse (\%) & Schema (\%) \\
\midrule
Gemini 3.1 Pro        & 87.5 & 97.3 \\
Gemini 3 Flash        & 83.3 & 97.3 \\
Gemma 4 31B           & 83.3 & 82.4 \\
Gemma 4 26B           & 79.2 & 97.3 \\
Gemini 2.5 Flash Lite & 41.7 & 93.2 \\
\bottomrule
\end{tabular}
\end{table}

Four of the five models achieve 79\% or higher on both categories. Flash Lite is the exception, with a 51.5 percentage-point gap between its schema repair (93.2\%) and its parse repair (41.7\%). This category-shaped weakness is invisible in the blended overall of 80.6\% and illustrates why per-category reporting matters for deployment decisions.

Table~\ref{tab:attempts} reports the attempt distribution per corrector aggregated across all five models.

\begin{table}[h]
\centering
\caption{Attempt distribution per corrector on the stratified corpus ($n = 98$), aggregated across five Google-family models. DNC is did-not-converge (all three attempts failed).}
\label{tab:attempts}
\begin{tabular}{lrrrr}
\toprule
Corrector & 1st (\%) & 2nd (\%) & 3rd (\%) & DNC (\%) \\
\midrule
FullSchema    & 75.1 & 6.1 & 1.4 & 17.3 \\
BudgetSchema  & 73.9 & 6.7 & 1.0 & 18.4 \\
RAMPART       & 66.7 & 19.4 & 2.9 & 11.0 \\
Verbal        & 59.8 & 23.3 & 4.5 & 12.4 \\
Raw           & 49.8 & 19.0 & 6.3 & 24.9 \\
\bottomrule
\end{tabular}
\end{table}

FullSchema and BudgetSchema resolve queries on the first attempt most often (73-75\%) but have the highest DNC rates (17-18\%). RAMPART has a lower first-attempt rate (66.7\%) but the lowest DNC rate (11.0\%), meaning the refinement loop is most productive when paired with RAMPART. The pattern reflects a structural difference in how the correctors use the loop. FullSchema and BudgetSchema are bimodal, tending to either fix the query immediately or fail on all three attempts. RAMPART and Verbal produce more second-attempt recoveries (19-23\%) because the structured error feedback and the prior-attempt context give the model new information on each iteration.

On a pure-schema corpus ($n = 100$, zero parse errors), overall success rates ranged from 86\% to 99\% with the RAMPART corrector, with first-attempt repair rates between 28\% and 81\%. The refinement loop recovered 18 to 61 percentage points on this harder corpus, confirming its contribution is robust to corpus composition.

\subsection{End-to-end evaluation on CypherBench}
\label{sec:results-feng2025cypherbench}

To verify that the validation and correction pipeline does not degrade queries that are already correct, we evaluated CYGNET end-to-end on seven test-split schemas from the CypherBench benchmark \citep{feng2025cypherbench}, totalling 2348 questions across five Google-family models. Each model generated Cypher from the natural-language question using CypherBench's own zero-shot prompt template. The generated query was then passed through CYGNET's validator chain against a mirror built from the CypherBench schema, corrected by the RAMPART corrector (up to 3 attempts) if validation failed, and executed against the CypherBench database. Execution accuracy (EX) was measured using CypherBench's evaluation harness, where EX=1 if the query's result set matches the ground truth.

\begin{table}[h]
\centering
\caption{CypherBench end-to-end evaluation (2348 questions per model, 7 schemas). One-shot is the generation baseline without CYGNET. Corrected is the result after CYGNET validation and correction. The structural-error fraction is the percentage of one-shot queries that fail to execute (broken Cypher), which is the upper bound on what CYGNET can catch.}
\label{tab:cypherbench-delta}
\small
\begin{tabular}{lrrrr}
\toprule
Model & One-shot EX (\%) & Corrected EX (\%) & $\Delta$ (pp) & Structural errors (\%) \\
\midrule
Gemini 3.1 Pro        & 85.3 & 85.2 & $-$0.01 & 0.6 \\
Gemini 3 Flash        & 83.8 & 85.2 & +1.35 & 3.9 \\
Gemma 4 31B           & 77.6 & 77.7 & +0.16 & 4.8 \\
Gemma 4 26B           & 64.3 & 64.4 & +0.04 & 19.3 \\
Gemini 2.5 Flash Lite & 43.8 & 44.5 & +0.65 & 7.1 \\
\bottomrule
\end{tabular}
\end{table}

CYGNET does not degrade generation accuracy on any model (Table~\ref{tab:cypherbench-delta}). The correction delta ranges from $-$0.01 to +1.35 percentage points. Gemini 3 Flash receives the largest lift (+1.35 pp), tying Gemini 3.1 Pro after correction at 85.2\%. On Pro itself the delta is net-flat because the model already produces structurally valid Cypher 99.4\% of the time, leaving almost no structural errors for the validator to catch.

The modest deltas are consistent with the structural-error fractions in the rightmost column. On the strongest models, fewer than 5\% of generated queries are structurally broken. The remaining failures (the gap between executable rate and execution accuracy) are semantic errors where the Cypher runs but returns wrong results. Structural validation cannot reach these. CypherBench reports executable rates ranging from 68.6\% (gemma2-9b) to 96.3\% (claude3.5-sonnet) on the same test split \citep{feng2025cypherbench}, confirming across model families that structural errors decline with model capability and the remaining accuracy gap is dominated by semantic failures that structural validation cannot reach.

\section{Prior art}
\label{sec:prior-art}

The gate we describe touches several areas of active research. Pre-execution validation of generated Cypher, schema filtering for query generation, Text-to-Cypher benchmarks and fine-tuning, synthetic graph construction for verification, cost gating for agent-generated queries, refinement loops with structured feedback from the SQL agent literature, schema introspection in knowledge-graph tooling, and multi-agent conversational systems over knowledge graphs all contribute pieces of the problem that the gate integrates. We position this work against each area in turn and summarise the feature comparison in Table~\ref{tab:prior-art}.

CyVer \citep{mandilara2025cyver} offers three validators (syntax, schema, property) operating in sequence with label inference for unlabelled nodes and scored validation results (weighted averages of correct nodes, relationships, and properties). The validators check queries against a provided schema description rather than against a runtime mirror, and the project does not include a cost gate, a refinement loop, or an agent-facing correction interface. Other Cypher parsers including \texttt{libcypher-parser} \citep{leishm2017libcypher} provide syntactic checking without schema awareness. LangChain's \texttt{CypherQueryCorrector} \citep{langchain2023corrector} repairs relationship-direction errors and ships as part of the standard LangChain Neo4j integration\footnote{The corrector addresses only relationship direction. Other forms of Cypher repair are not part of its scope.}. Recent work by Ozsoy \citep{ozsoy2026confidence} extends confidence-based Text2Cypher inference with grammar validation and schema-constraint filtering applied as a post-generation filter. That work focuses on selecting the best candidate from multiple generations rather than iterating a correction loop, and reports that schema-aware filtering improves execution quality but increases the number of empty predictions.

The role of schema filtering in query generation is investigated separately by Ozsoy \citep{ozsoy2025schemafilter}, who reports that filtering the schema to relevant elements improves query generation for smaller models while larger models benefit less due to their longer context capabilities. Schema filtering also reduces token cost for all model sizes. These findings are consistent with the CYGNET corrector comparison, where the FullSchema corrector (which passes the complete schema) matches the RAMPART corrector on Gemini 3.1 Pro but underperforms on Gemini 2.5 Flash Lite. The schema-filtering principle that Ozsoy identifies for the generation task transfers to the correction task, with the additional observation that on some models the filtered prompt can hurt parse-error repair when the filtering constrains the response space.

Three benchmark efforts define the current evaluation landscape for Cypher generation. The Neo4j Text2Cypher dataset \citep{ozsoy2025text2cypher} aggregates 44387 instances from publicly available sources with a training and test split, reporting that fine-tuned models consistently improve over their baselines. CypherBench \citep{feng2025cypherbench} introduces 11 Wikidata-derived property graphs with over 10000 questions spanning 12 graph matching patterns, reporting GPT-4o at 60.18\% execution accuracy and no open-weights model under 10B parameters exceeding 20\%. SynthCypher \citep{tiwari2024synthcypher} generates 29800 synthetic Text-to-Cypher instances and demonstrates up to 40\% absolute improvement from fine-tuning on open-weights models. None of the three benchmarks decompose their headline accuracy into format-compliance failures and wrong-Cypher failures. All three use model-agnostic prompt templates without per-model tuning. A separate line of work applies reinforcement learning with Group Relative Policy Optimisation to refine small language models for Text2Cypher beyond what supervised fine-tuning achieves \citep{tran2025refining}. Multilingual evaluation of Text2Cypher models \citep{ozsoy2025multilingual} demonstrates that the task extends beyond English with varying performance across languages.

SynthCypher and SyntheT2C \citep{zhong2025synthet2c} both use synthetic graph construction for offline verification of generated Cypher. SynthCypher populates Neo4j instances with LLM-generated data. SyntheT2C validates against existing medical knowledge graph databases. Both use execution-based verification as part of a training data pipeline rather than at runtime in an agent loop. CYGNET adopts the synthetic graph idea but uses a minimal one-node-per-label construction at runtime as a planning and execution target. No published system uses a constructed mirror graph for runtime per-query validation.

Cost-aware querying for agent-generated queries is an active research area in the SQL community. Recent work on cost-aware text-to-SQL \citep{deochake2025costaware} reports cost variance of more than three times between LLMs on identical questions and proposes cost-awareness as an evaluation dimension absent from established benchmarks. BIRD \citep{li2023bird} introduces a Valid Efficiency Score that combines correctness with wall-clock execution time, although the metric does not capture cloud compute cost directly. Pre-execution cost estimation for non-SQL query languages appears in research on GraphQL \citep{mavroudeas2021graphql} where machine learning models trained on query-level features (field counts, AST structure, static analysis upper bounds) estimate query expense before execution. Production Cypher tooling such as Neo4j Labs' \texttt{mcp-neo4j-cypher} \citep{neo4j2024mcp} addresses runaway queries through post-hoc timeouts and truncation rather than pre-execution estimation. CYGNET applies the cost-aware pattern from the SQL agent literature to Cypher with the planner's own EXPLAIN output as the cost signal.

The pattern of using execution feedback to refine LLM-generated queries is mature in the SQL agent literature. MAC-SQL \citep{wang2023macsql} introduces a Refiner agent that diagnoses syntactic correctness, execution feasibility, and result emptiness, applying corrections iteratively. DIN-SQL \citep{pourreza2023dinsql} introduces decomposed in-context learning with self-correction, though the correction step does not consume specific error information. RetrySQL \citep{retrysql} trains text-to-SQL models on retry data containing both incorrect and corrected steps, demonstrating that self-correction can be learned through continued pre-training and that the resulting models are competitive with proprietary models of much larger size. Self-correction distillation \citep{selfcorrection_distillation} classifies LLM-generated query errors into types and provides type-specific feedback in a closed-loop mechanism, then distils both the generation and error-correction capabilities from a teacher model to a smaller student. The follow-on text-to-SQL literature converges on execution-guided refinement \citep{wang2018executionguided,ni2023lever,chen2023selfdebug}, with surveys identifying error analysis as a substantive component of overall system quality \citep{hong2025text2sql_survey,shi2025text2sql_survey,singh2025text2sql_survey}. The pattern had not been packaged with a controlled error vocabulary and a registry-based prompt assembler for Cypher before CYGNET.

Schema introspection as a first-class agent step is standard practice across knowledge-graph tooling. Neo4j Labs' \texttt{mcp-neo4j-cypher} server \citep{neo4j2024mcp} exposes schema introspection alongside read and write tools, requiring the APOC plugin for schema inspection. LangChain's \texttt{GraphCypherQAChain} \citep{langchain2023corrector} and the LlamaIndex knowledge-graph query engine \citep{llamaindex2024kg} inject the schema into the LLM's prompt at query-generation time.

Multi-agent conversational systems over knowledge graphs are a recent and active area. AGENTiGraph \citep{zhao2025agentigraph} demonstrates intent classification, task planning, and automated knowledge integration over Neo4j, with its Knowledge Graph Interaction Agent translating user intents into executable Cypher queries and its Dynamic Knowledge Integration Agent using LLM-generated Cypher to update the graph structure. The system reports 90.45\% execution success rate on a 3500-query test set, but includes no pre-execution structural validation of the generated Cypher. Chatty-KG \citep{omar2025chattykg} combines retrieval-augmented generation with structured SPARQL query generation through task-specialised LLM agents, achieving up to 83.3\% P@1 on domain-specific knowledge graphs. The system applies assertion-based validation at multiple pipeline stages (triples, vertices, and query structure) but these checks validate intermediate representations and SPARQL syntax rather than query structure against the graph schema through a mirror database. KG-RAG \citep{zhu2025kg2rag} extends the retrieval-augmented generation pattern with explicit graph structure for chunk relationships. GraphSeek \citep{besta2026graphseek} replaces direct query generation from natural language with planning over a Semantic Catalog that describes both the graph schema (through concise natural-language annotations on node and relationship types) and the available graph operations, separating an LLM planning plane from a deterministic non-LLM execution plane and achieving 86\% success rate over enhanced LangChain on complex graph analytics tasks. None of these systems include a pre-execution structural validation layer that checks generated queries against the graph schema before they reach production.

Table~\ref{tab:prior-art} summarises the comparison across differentiating features. CyVer v2 narrows the validation gap with syntax, schema, and property validators plus scored validation results, but does not provide cost gating, a refinement loop, mirror-graph execution, or MCP transport. AGENTiGraph generates Cypher for Neo4j through LLM agents but provides no structural validation, cost gating, or error vocabulary, and its 9.55\% execution failure rate (on a 3500-query test set) includes structural errors that a pre-execution gate could catch. GraphSeek addresses the graph-analytics problem with a Semantic Catalog abstraction but operates at a different level, providing multi-step analytics planning with self-correction through its Decision stage rather than per-query structural validation. The Text-to-Cypher benchmarks and fine-tuning work establish the evaluation landscape but do not package a runtime validation or correction system.

\begin{landscape}
\begin{table}
\centering
\begin{threeparttable}
\caption{Comparison of CYGNET against the most directly comparable Cypher and graph-query tooling on differentiating features. Y indicates full support, N indicates absent, P indicates partial support with the limitation noted in the table notes.}
\label{tab:prior-art}
\begin{tabularx}{\linewidth}{X c c c c c c c c}
\toprule
Feature & CyVer v2\tnote{j} & LangChain CQC\tnote{a} & mcp-neo4j-cypher\tnote{k} & GraphSeek\tnote{b} & MAC-SQL\tnote{c} & Ozsoy 2026\tnote{d} & AGENTiGraph\tnote{e} & CYGNET \\
\midrule
Parse validation                & Y & P\tnote{f} & N & N & Y & Y & N & Y \\
Schema-spec validation          & Y & P\tnote{f} & N & N & N/A & P\tnote{g} & N & Y \\
Property validation             & Y & N & N & N & N/A & N & N & Y \\
Schema introspection            & Y & Y & Y & Y & N/A & N & Y & Y \\
EXPLAIN-based cost gating       & N & N & N & N & N\tnote{h} & N & N & Y \\
Structured error vocabulary     & P\tnote{i} & N & N & N & Y & N & N & Y \\
Agent-facing refinement loop    & N & N & N & N & Y & N & N & Y \\
Schema filtering                & N & N & N & Y & Y & Y & N & Y \\
MCP transport                   & N & N & Y & N & N & N & N & Y \\
Mirror-graph execution          & N & N & N & N & N & N & N & Y \\
\bottomrule
\end{tabularx}
\begin{tablenotes}\footnotesize
\item[a] LangChain CypherQueryCorrector \citep{langchain2023corrector}.
\item[b] Operates on multi-step analytics planning \citep{besta2026graphseek}, not per-query validation.
\item[c] For reference, applies to SQL rather than Cypher \citep{wang2023macsql}.
\item[d] Confidence-based filtering with grammar and schema constraints \citep{ozsoy2026confidence}.
\item[e] Generates Cypher for Neo4j via LLM agents \citep{zhao2025agentigraph}. Reports 90.45\% execution success with no pre-execution query validation.
\item[f] Limited to relationship-direction repair.
\item[g] Relationship-direction checking only.
\item[h] Post-hoc execution cost only (Valid Efficiency Score in BIRD benchmark).
\item[i] Returns scored validation results, not categorized error payloads.
\item[j] Three validators (syntax, schema, property) with scored results \citep{mandilara2025cyver}.
\item[k] Neo4j Labs MCP server exposing schema introspection and query execution \citep{neo4j2024mcp}.
\end{tablenotes}
\end{threeparttable}
\end{table}
\end{landscape}

\section{Conclusions}
\label{sec:conclusions}

We presented CYGNET, a pre-execution validation and correction system that gates agent-generated Cypher queries before they reach a Neo4j production database. It validates generated Cypher against the graph schema without touching the production database. It estimates query cost and flags expensive queries before they execute. It corrects broken queries by feeding structured error descriptions to a language model and iterating until the query passes validation or an attempt budget is exhausted. The system combines a four-backend validator chain operating in short-circuit or collect-all modes, an EXPLAIN-based cost gate with operator-aware mitigation suggestions, a structured five-category error vocabulary, and five pluggable correctors implementing a single-shot protocol with a generic refinement loop. On a template-generated corpus covering four error categories across nine schemas, the gate catches 100\% of parse errors, 100\% of constraint violations, and 100\% of schema-reference errors in path queries with labelled endpoints, at a false-positive rate of zero across 1135 queries. The EXPLAIN-based cost gate identifies catastrophic plan structures such as unbounded variable-length paths and full-scan aggregations that would otherwise consume production resources. Planner row estimates are well-calibrated on five of seven tested query patterns (overall median ratio 1.0) and serve as conservative risk signals, with known overestimation on variable-length unbounded paths and property-predicate selectivity where the planner lacks cardinality statistics. Full-pipeline gate latency has a median of 5.6 ms, leaving substantial headroom under the 50 ms target for interactive agent use.

On a stratified corpus of 98 LLM-generated broken Cypher queries across five Google-family models, the RAMPART corrector achieves 80.6\% to 95.9\% overall success after up to three refinement attempts, with prompt-following rates between 88.0\% and 100.0\%. The refinement loop contributes 14 to 28 percentage points of the overall success on this corpus, recovering queries that the first attempt failed to fix. The corrector comparison shows that no single corrector dominates across all models, and the per-corrector gap is driven primarily by prompt following rather than Cypher repair quality. The RAMPART corrector's 17-fold token cost over the Raw corrector is justified on models that need the additional structure to follow the response protocol (Gemma 4 31B gains 36 percentage points in prompt following) and is unnecessary overhead on models that comply without it (Gemini 3.1 Pro achieves the same success rate on either). Prompt strategy is model-dependent in practice, and all five correctors are available so users can match the prompt complexity to their model's capabilities and their token budget. Pipeline engineering across the schema-faithful mirror, response-format specification, retry policy, and prompt design lifted overall success by 21 to 66 percentage points compared to the initial infrastructure, demonstrating that the system's value depends on careful optimisation of the entire call chain rather than on any single component. No validation divergences were observed between YAML-built mirror, introspection-built mirror, and source database across ten databases of varying schema shapes on every query in the test sample.

On seven schemas from the CypherBench benchmark (2348 questions, ACL 2025), the validation and correction pipeline does not degrade generation accuracy on any of the five Google-family models tested. Correction deltas range from $-$0.01 to +1.35 percentage points. The modest deltas reflect the fact that on the strongest models fewer than 5\% of generated queries are structurally broken, leaving little for the structural validator to catch. The system operates as a safe defensive layer that catches and corrects structural errors when they occur without introducing regressions on queries that are already correct.

The validator's 0\% recall on property sibling swaps (where an agent substitutes a property name that is valid in the schema but appears on a different label than the query's binding context requires) marks a formal boundary of structural validation. Any property name that exists anywhere in the schema will pass structural checks regardless of whether it is correct for the label in question. Detecting such errors requires either type-aware property resolution that tracks which properties are declared on which labels through the query's variable bindings, or semantic validation against user intent. The first is an engineering extension within the current architecture, the second is a different problem.

The AST backend uses a Cypher 9 era ANTLR grammar that does not parse the CALL subquery syntax introduced in Neo4j 4.0. Queries using this syntax produce a parse error that the chain's later backends compensate for. The Schema model represents a single source-target label pair per relationship type, which collapses multi-endpoint relationships observed in production graphs to the first observed pair. Multi-label co-occurrence groups are not represented. The nine-schema test corpus exercises four error categories but uses template-generated queries rather than real LLM output.

\bibliographystyle{unsrtnat}
\bibliography{writeup}

\appendix

\section{Corrector prompt examples}
\label{app:prompts}

The following listings show the user prompt each corrector sends to the language model for the query \texttt{MATCH (n:Moive) RETURN n.title} against the recommendations schema (a schema error where \texttt{Moive} should be \texttt{Movie}). All five correctors share the same system prompt (613 characters). The system prompt instructs the model to return a JSON object with a \texttt{cypher} field and includes a one-shot correction example.

\subsection*{Raw corrector (492 characters)}

The Raw corrector passes a flat schema summary (labels and relationship types, one line each) and the Neo4j error string as-is.

\begin{lstlisting}[basicstyle=\ttfamily\footnotesize,breaklines=true]
## Schema

Labels: Movie, Person, Genre, User, Actor, Director
Relationships:
  - (:Person)-[:ACTED_IN]->(:Movie)
  - (:Person)-[:DIRECTED]->(:Movie)
  - (:Person)-[:PRODUCED]->(:Movie)
  - (:Person)-[:WROTE]->(:Movie)
  - (:Movie)-[:IN_GENRE]->(:Genre)
  - (:User)-[:RATED]->(:Movie)

## Broken query

MATCH (n:Moive) RETURN n.title

## Error from Neo4j

The label `Moive` does not exist.

Return a JSON object: {"cypher": "<corrected query>"}.
\end{lstlisting}

\subsection*{Verbal corrector (652 characters)}

The Verbal corrector uses the same flat schema summary but renders the structured error payload as a prose paragraph with did-you-mean suggestions.

\begin{lstlisting}[basicstyle=\ttfamily\footnotesize,breaklines=true]
## Schema

Labels: Movie, Person, Genre, User, Actor, Director
Relationships:
  - (:Person)-[:ACTED_IN]->(:Movie)
  - (:Person)-[:DIRECTED]->(:Movie)
  - (:Person)-[:PRODUCED]->(:Movie)
  - (:Person)-[:WROTE]->(:Movie)
  - (:Movie)-[:IN_GENRE]->(:Genre)
  - (:User)-[:RATED]->(:Movie)

## Broken query

MATCH (n:Moive) RETURN n.title

## Why the query is broken

The label `Moive` referenced in the query is not declared
in the schema. Did you mean one of: `Movie`? Other valid
labels in scope include: `Actor`, `Director`, `Genre`,
`Movie`, `Person`, `User`.

Return a JSON object: {"cypher": "<corrected query>"}.
\end{lstlisting}

\subsection*{FullSchema corrector (1830 characters)}

The FullSchema corrector passes the complete schema (every label with all properties and types, every relationship type with source and target labels, all constraints) plus the structured error as a JSON payload.

\begin{lstlisting}[basicstyle=\ttfamily\footnotesize,breaklines=true]
## Schema

### Labels
- Actor: bio:STRING?, born:DATE?, bornIn:STRING?,
  died:DATE?, name:STRING, poster:STRING?, tmdbId:STRING,
  url:STRING?
- Director: (same property set as Actor)
- Genre: name:STRING
- Movie: budget:INTEGER?, countries:LIST?,
  imdbId:STRING?, imdbRating:FLOAT?, imdbVotes:INTEGER?,
  languages:LIST?, movieId:STRING, plot:STRING?,
  poster:STRING?, released:INTEGER?, revenue:INTEGER?,
  runtime:INTEGER?, tagline:STRING?, title:STRING,
  url:STRING?, year:INTEGER?
- Person: (same property set as Actor)
- User: name:STRING?, userId:STRING

### Relationships
- (:Person)-[:ACTED_IN]->(:Movie) props=[role:STRING?]
- (:Person)-[:DIRECTED]->(:Movie)
- (:Movie)-[:IN_GENRE]->(:Genre)
- (:Person)-[:PRODUCED]->(:Movie)
- (:User)-[:RATED]->(:Movie) props=[rating:FLOAT,
  timestamp:INTEGER?]
- (:Person)-[:WROTE]->(:Movie)

### Constraints
- genre_name_unique: UNIQUENESS on :Genre(name)
- movie_movieId_unique: UNIQUENESS on :Movie(movieId)
- person_tmdbId_unique: UNIQUENESS on :Person(tmdbId)
- user_userId_unique: UNIQUENESS on :User(userId)

## Failing query

MATCH (n:Moive) RETURN n.title

## Error payload (JSON)

{"category":"schema", "unknown_reference":"Moive",
 "reference_kind":"label",
 "did_you_mean":["Movie"],
 "available_in_scope":["Actor","Director","Genre",
   "Movie","Person","User"]}

Return a JSON object: {"cypher": "<corrected query>"}.
\end{lstlisting}

\subsection*{RAMPART corrector (8613 characters, abbreviated)}

The RAMPART corrector assembles its prompt from stored prompt blocks. The compiled prompt for this query includes a category-matched error-vocabulary block (schema-error refinement guidance with instructions on how to use the \texttt{did\_you\_mean} and \texttt{available\_in\_scope} fields), per-label schema blocks filtered to the labels the query references (Movie and its neighbours), pattern-hint blocks (simple MATCH, aggregation, relationship traversal, OPTIONAL MATCH, WITH staging), and the structured error payload. The token budget removes lower-priority blocks when the prompt exceeds the configured limit.

\begin{lstlisting}[basicstyle=\ttfamily\footnotesize,breaklines=true]
# Schema-error refinement

A ``SchemaError'' means the query referenced a label,
relationship type, or property that is not declared in the
schema. The structural validator has identified exactly
which token is unknown and what kind of reference it is.

## Key fields

- ``unknown_reference'': the token that does not match
  any declared name.
- ``reference_kind'': one of ``label'', ``relationship'',
  or ``property''.
- ``did_you_mean'': pre-ranked close matches. Prefer these.
- ``available_in_scope'': the full option space at the
  failure site.

## Typical refinements

1. did_you_mean has one strong match -- substitute and return
2. did_you_mean is empty but available_in_scope is short --
   pick the most semantically plausible option
3. available_in_scope_truncated is True -- did_you_mean is
   the only hint; if empty, abort

---
# Refinement intent (attempt 1)

## Failing query
MATCH (n:Moive) RETURN n.title

## Failure category: schema
## Error payload (JSON)
{"category":"schema", "unknown_reference":"Moive", ...}

---
# Schema: label Movie
- movieId: STRING (required)
- title: STRING (required)
[... 14 more properties ...]

# Schema: label Person
[... 8 properties ...]

[... 4 more label blocks, 6 relationship blocks,
 5 pattern-hint blocks ...]
\end{lstlisting}

The RAMPART prompt is 17 times larger than the Raw prompt (8613 vs 492 characters). On Gemini 3.1 Pro the additional context does not improve correction success (both achieve 93-95\% overall). On Gemma 4 31B the structured error-vocabulary guidance and the did-you-mean fields lift prompt-following from 53\% (Raw) to 89\% (RAMPART), a 36 percentage-point gain. The token cost is justified on models that benefit from the additional structure and is unnecessary overhead on models that do not.

\section{Error vocabulary examples}
\label{app:errors}

The following JSON payloads show the structured error each category produces for a representative query against the recommendations schema.

\subsection*{Parse error}

Triggered by \texttt{MATCH n:Movie RETURN n} (missing parentheses around the node pattern).

\begin{lstlisting}[basicstyle=\ttfamily\footnotesize,breaklines=true]
{
  "category": "parse",
  "message": "no viable alternative at input 'MATCH n:'",
  "line": 1,
  "column": 8,
  "snippet": "MATCH n:Movie RETURN n",
  "excerpt_with_caret": "MATCH n:Movie RETURN n\n       ^"
}
\end{lstlisting}

\subsection*{Schema error}

Triggered by \texttt{MATCH (n:Moive) RETURN n.title} (misspelled label).

\begin{lstlisting}[basicstyle=\ttfamily\footnotesize,breaklines=true]
{
  "category": "schema",
  "unknown_reference": "Moive",
  "reference_kind": "label",
  "did_you_mean": ["Movie"],
  "query_context": "MATCH (n:Moive) RETURN n.title",
  "available_in_scope": ["Actor", "Director", "Genre",
    "Movie", "Person", "User"],
  "available_in_scope_truncated": false
}
\end{lstlisting}

\subsection*{Property error}

Triggered by \texttt{MATCH (n:Movie) WHERE n.released = 'not-an-int' RETURN n} (string literal for an INTEGER property).

\begin{lstlisting}[basicstyle=\ttfamily\footnotesize,breaklines=true]
{
  "category": "property",
  "property_name": "released",
  "declared_type": "INTEGER",
  "used_type": "STRING",
  "query_context": "MATCH (n:Movie) WHERE n.released = ...",
  "did_you_mean": []
}
\end{lstlisting}

\subsection*{Constraint error}

Triggered by \texttt{CREATE (m:Movie) RETURN m} (missing the required \texttt{movieId} property declared by a uniqueness constraint).

\begin{lstlisting}[basicstyle=\ttfamily\footnotesize,breaklines=true]
{
  "category": "constraint",
  "constraint_id": "_implicit_Movie_movieId_required",
  "constraint_kind": "existence",
  "property_name": "movieId",
  "violating_value": null
}
\end{lstlisting}

Each payload carries enough information for a downstream corrector to attempt a targeted fix without re-examining the schema. The \texttt{did\_you\_mean} field in the schema-error payload provides edit-distance-ranked alternatives. The \texttt{declared\_type} and \texttt{used\_type} fields in the property-error payload identify the exact mismatch. The \texttt{constraint\_kind} field in the constraint-error payload distinguishes uniqueness from existence violations. The corrector uses these fields to construct a focused prompt rather than re-stating the entire schema.

\end{document}